\documentclass[a4paper]{itatnew}
\usepackage{cite,amssymb,amsfonts,algorithm,algpseudocode,graphicx,xcolor,rotating,multirow,array,balance} 
\usepackage[centertags]{amsmath}
\extrafloats{100}

\newcounter{enumroman}

\title{Using Artificial Neural Networks to Determine Ontologies Most Relevant to Scientific Texts}
\author{Luk\'a\v{s} Korel\inst{1} \and Alexander S. Behr\inst{2} \and Norbert Kockmann\inst{2} \and Martin Hole\v{n}a\inst{1,3,4}}

\institute{Faculty of Information Technology, CTU, Prague, Czech Republic 
\and 
Faculty of Biochemical and Chemical Engineering, TU Dortmund University, Germany, \email{\{alexander.behr|norbert.kockmann\}@tu-dortmund.de}
\and 
Institute of Computer Science, Czech Academy of Sciences, Prague, \email{martin@cs.cas.cz}
\and 
Leibniz Institute for Catalysis, Rostock, Germany, \email{martin.holena@catalysis.de}}

\begin{document}
\bibliographystyle{plain}

\twocolumn[\maketitle]

\begin{abstract}
This paper provides an insight into the possibility of how to find ontologies most relevant to scientific texts using artificial neural networks. The basic idea of the presented approach is to select a representative paragraph from a source text file, embed it to a vector space by a pre-trained fine-tuned transformer, and classify the embedded vector according to its relevance to a target ontology. We have considered different classifiers to categorize the output from the transformer, in particular random forest, support vector machine, multilayer perceptron, k-nearest neighbors, and Gaussian process classifiers. Their suitability has been evaluated in a use case with ontologies and scientific texts concerning catalysis research. The obtained results confirm support vector machines as a promising classifier, but surprisingly show random forest as unsuitable for this task.

\end{abstract}

\section{Introduction}
A domain ontology defines a set of representational primitives with which to model a domain of knowledge or discourse. The representational primitives are typically classes, attributes, and relationships. The definitions of the representational primitives include information about their meaning and constraints on their logically consistent application. Classes can be defined in two ways: by annotating their definitions, or by connecting classes with each other and with properties. Each domain ontology typically uses domain-specific definitions of terms denoting its primitives.

The FAIR research data management (Findable, Accessable, Interoperable, and Reuseable) needs a consistent data representation in ontologies, particularly for representing the data structure in the specific domain \cite{wulf21unified}. Since different ontologies are written by different people, they are often incompatible, even within the same domain. As systems that rely on domain ontologies expand, it is often needed to merge domain ontologies by manual tuning. The same is true for enhancing an ontology with information available in domain-related texts. Merging and enhancing ontologies is thus a largely manual process and therefore time-consuming and expensive.

The need to find a suitable ontology for an input text can help in classifying the information presented within the text as well as to connect the input text with data. This would allow for automated selection of ontologies and respective classification of the text. Different text data could thus be compared automatically in an understandable way and connected with corresponding research data. Ontologies represent "a formal specification of a shared conceptualization" \cite{C3} and can thus be used to express knowledge and data in a formalized, standardized description language to specify terms and relations between those terms.

Current ontology recommenders, such as the NCBO ontology recommender \cite{C4}, score annotations based on words similar to preferred and alternate labels of ontology classes and term frequency. In contrast to this, this work aims to use text representation learning in order to not only search for words also contained in ontologies but also to find concepts with similar semantic meaning between text and ontology.

This paper is devoted to a specific problem encountered during enhancing ontologies and sometimes during their merging: to decide which of several available ontologies is most relevant to given domain-related piece of text. Our solution to the problem relies primarily on artificial neural networks (ANNs), in particular on natural language processing (NLP).

The next section surveys the applicability of artificial neural networks to ontologies. Section \ref{Methodological} recalls the employed methods of text preprocessing. There have been used modules for text extractions from PDF files, for transforming extracted files to pure text and for eliminating irrelevant paragraphs. In the section, text representation learning is described as well as the principles of the employed classifiers. In section \ref{Case}, an application of the proposed methodology to catalysis is described and evaluated.

With regard to sources we have studied described in part \ref{Applicability} of this article, we are not aware that classifiers learned from the results of representational learning have ever been used to determine the most relevant of a given set of ontologies.

\section{Applicability of Artificial Neural Networks to Ontologies}
\label{Applicability}

In connection with learning and extending ontologies, artificial neural networks (ANNs) have been primarily used for identification of concepts, relations and attributes \cite{3, 18, 28}. With respect to relations, some ANN-based methods have been developed specifically for subsumption relations needed for the construction of taxonomies \cite{5, 14, 38, 57}. In connection with integration of ontologies, they have been primarily used for ontologies matching aka ontologies alignment \cite{7,8,20,67}. The variety of employed kinds of ANNs is rather large. It includes traditional multilayer perceptrons (MLPs) \cite{31}, adaptive resonance theory (ART) networks \cite{27} and associative memories \cite{40}, as well as the modern deep convolutional networks (CNNs) \cite{7,36}, deep belief networks \cite{3}, long short term memory (LSTM) networks together with their bidirectional variant (BiLSTM) \cite{42} and gated recurrent units (GRU) networks \cite{52,53}. The dependence of ontologies on texts led to using networks developed for text and natural language representation learning, most importantly BERT \cite{39, 48}, the bidirectional encoder representations from transformers, and word2vec \cite{43}, the most traditional network for embedding text into an Euclidean space. The close relationship of ontologies to knowledge graphs led to using also RDF2Vec \cite{38, 53}, which was originally proposed for knowledge graphs \cite{58}. In connection with word2vec and RDF2Vec, it is on similar principles, the network OWL2Vec was proposed for embedding of ontologies \cite{59}. Finally, the graph-like structure of ontologies brought usage graph neural networks (GNNs) \cite{20,67}.

Closest to the proposed project is the way ANNs have been recently used in connection with translating into OWL \cite{51,52}, with predicate chaining and restriction \cite{40}, and with taxonomy extraction from knowledge graphs \cite{38}. In \cite{51}, ontology learning is tailored as a transductive reasoning task that uses two recurrent neural networks to translate text in natural language into OWL specifications in description logic. That approach was further developed in \cite{52}, resulting in a system based on a single recurrent network of GRU type. It uses an encoder-decoder configuration and translates through syntactic transformation a subset of natural language into the description logic language ALLQ. Moreover, the system generalizes over different syntactic structures, and has the ability to tolerate unknown words through copying input words as extralogical symbols to the output, as well as the ability to enrich the training set with new annotated examples. In \cite{40}, a mapping is established between ontologies and a pair of interacting associative memories. One of them stores assertions, and the other stores entailment rules. The most recent work \cite{38} describes a method for the specific task of extracting a taxonomy from an embedding of a knowledge graph. Over that embedding, which can be obtained for example with RDF2Vec, hierarchical agglomerative clustering is performed, first without using type information, and then injecting types into the hierarchical clustering tree. In addition, an axiom induction algorithm is applied to each cluster in the resulting tree, which allows to identify new classes corresponding to those axioms that describe their respective clusters accurately enough.

Neural networks are often used due to their strengths in natural language processing task. Ontology construction rely very much on texts, which suggest the applicability of artificial neural networks (ANNs) in this context.

\section{Methodological Background}
\label{Methodological}
This section describes details of used methods. In the first part, text pre-processing is sketched. The second part describes using a transformer for embedding input paragraphs to classification numeric vectors. The final part recalls used classifiers, which use outputs from the transformer for classification to an available ontology.

\subsection{Text Preprocessing}
For the problem of scientific texts classification to the most relevant existing ontology, we have been using documents in portable document files (PDFs). An issue with PDFs is that they are optimized to print on physical printer, thus they contain meta-information about the contained text related to the position on the page. Therefore, it is not easy to address a single paragraph. If the file is read using the basic library for PDF files and the newline mark is used as the splitter, it returns only a single row, not the whole paragraph. Another issue is connected with multi-column documents. If the document does not include information about where the text continues, software libraries for text extraction from PDF usually continue with the next letter on the same row.

One solution to get text data from multi-column PDF is to use Microsoft Word engine. Its engine is able to solve both problems and parse text properly. It identifies structural information in text such as headings, paragraphs and sentences. Each document may contain texts irrelevant to the topic of interest, for example references, acknowledgement etc.

Specifications of the ontologies are most often stored in OWL files. OWL \cite{OWL} is a specific kind of XML for ontologies. Text that describes classes and relations may be stored in different tags, depending on the decision of the ontology designer.

\subsection{Text Representation Learning}
For typical data analysis tasks like classification of clustering, it is suitable to represent words or other parts of text by vectors in an Euclidean space. Such representation is mostly the result of representation learning by ANNs. In the area of text analysis and processing, the probably most successful representation learning algorithm is BERT (Bidirectional Encoder Representations from Transformers)\cite{BERT}.

BERT needs to be trained using large amount of texts. That is why some pretrained version is typically used, and often subsequently fine-tuned using texts concerning the considered topic. Such fine-tuning is often performed even if the pretrained network was trained, apart from general texts, also with texts from some broader relevant domain (biology, medicine, chemistry, etc.).

\begin{figure}[htp]
  \centering
    \includegraphics[width=0.48\textwidth]{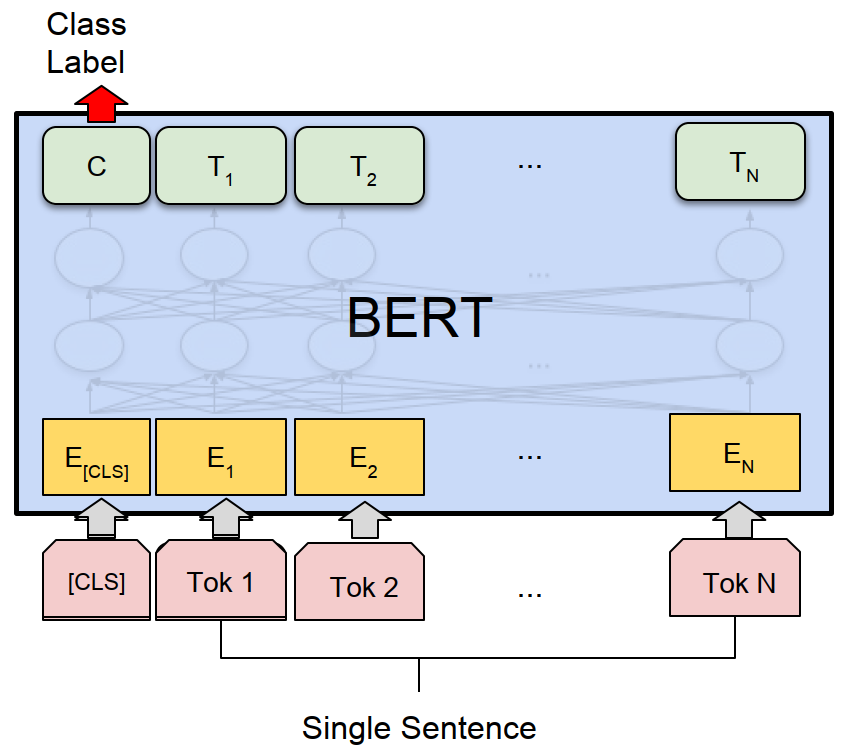}
    \caption{BERT (Bidirectional Encoder Representations from Transformers) architecture \cite{BERT}. An input sentence is divided into tokens and BERT represents each token  with one numeric vector, taking into account the predecessors of that token in the sentence. The vector C, which represents the token CLS closing the sentence, is used for classification.}
    \label{BERT}
\end{figure}

The basic schema of BERT is given in Figure \ref{BERT}. The tokenized input at first passes through the encoder, which embeds sentences to elements of an Euclidean space. These vectors are used as input to the BERT decoder. BERT returns one vector for each input. Each input sequence contains a special token at the beginning marked as CLS. Vectors embedding the tokens of an input sequence can be arranged into a matrix. The first row of the matrix is the embedding of the whole input. Details of BERT are described in \cite{BERT} and on the webpage \href{https://huggingface.co/docs/transformers/model_doc/bert}{https://huggingface.co/docs/transformers/model\_doc/bert}. These embeddings of every input paragraph are taken into account for the final assignment of the most relevant ontology to the paragraph.

\subsection{Classification}
The embeddings obtained in BERT are used as inputs for classifiers classifying a given input part of text (e.g., a paragraph) with respect to its relevance to the considered ontologies. Those classifiers have been trained on the embeddings of the annotations from the considered ontologies because for them, the ground truth (i.e., the ontology to which the annotation belongs) is known.

We have decided to select five classifiers implemented in the python library scikit-learn \cite{sklearn}. They are the following:
\begin{enumerate}
  \item Random forest (RF): An ensemble classifier that fits a number of classification trees on various sub-samples of the training data and uses some aggregation function to improve the predictive accuracy and control over-fitting. Usually, each tree in the ensemble is built using a sample drawn with replacement (i.e., a bootstrap sample) from the training set. Furthermore, when splitting each node during the construction of a tree, the best split is found using either all input features or a random subset of a given size. The purpose is to decrease the variance of the forest estimator. Indeed, individual decision trees typically exhibit high variance and tend to overfit. The injected randomness in forests yield decision trees with somewhat decoupled prediction errors. By taking an average of those predictions, some errors can cancel out. RFs achieve a reduced variance by combining diverse trees, sometimes at the cost of a slight increase in bias. Typically, the variance reduction yields an overall better model \cite{RDF}.
  \item Support vector machine (SVM): It is a classifier designed specifically to achieve the lowest possible predictive error, using a known relationship between generalization error and margin of the separating hyperplane. It uses only training points on both support hyperplanes of the margin (support vectors), so it is also memory efficient. A simple SVM can be used only for linearly separable classes. For linearly nonseparable classes, the data must be first transformed to linearly separable sets of functions  in a high-dimensional vector space of functions using a suitable kernel. The SVM classification has multiclass support handled according to a one-vs-one or one-vs-rest scheme \cite{SVM}.
  \item Gaussian Process (GP): It has been designed primarily for regression problems. A Gaussian Process Classifier (GPC) implements a collection of random variables indexed by an Eucliedan space for classification purposes through placing a GP prior on latent functions. Its purpose is to allow a convenient formulation of the classification through a logistic link function. GPCs support multi-class classification by performing either one-versus-rest or one-versus-one training and prediction. A crucial ingredient of each GPC is the covariance functions of the underlying GP. It encodes the assumptions on the similarity of Gaussian distributions corresponding to different points \cite{GPC}.
  \item K nearest neighbors: Neighbors-based classification simply stores instances of the training data. A query point is assigned the data class which has the most representatives within the nearest neighbors of the point. The nearest neighbors classification can use uniform weights, that means, the value assigned to a query point is computed from a simple majority vote of the nearest neighbors. In some cases, it is better to weight the neighbors in such a way that nearer neighbors contribute more to the fit. For example, when an unknown point's class is computed from two nearest neighbors and one of this two is nearer than other, it is useful to weight the class of the closer neighbor more than the class of the more distant one. The distance $d$ between two points can be computed as: $d(x,y)=(\sum_{i=1}^{n} |x_i - y_i|^c )^{\frac{1}{c}}$, where $n$ is the dimension of each point and $c \ge 1$, if $c = 1$, this is the Manhattan distance and in case $c = 2$, this is the Euclidean distance \cite{KNN}.
  \item Multi-layer Perceptron (MLP): Given a set of features  and a target, it can learn a non-linear function approximator for either classification or regression. It is different from logistic regression, because between the input and the output layer, there can be one or more non-linear hidden layers. The input layer consists of a set of neurons  representing the input features. Each neuron in the hidden layer transforms the values from the previous layer with a weighted linear summation, followed by a non-linear activation function. The output layer receives the values from the last hidden layer and transforms them into output values. The advantages of MLP are capability to learn non-linear models and capability to learn models in real-time (on-line learning). But the MLP with hidden layers have a non-convex loss function where there exists more than one local minimum. Therefore, different random weight initializations can lead to different validation accuracy. A MLP requires tuning a number of hyperparameters such as the number of hidden neurons, layers, and iterations. Moreover, it is sensitive to feature scaling \cite{MLP}.
\end{enumerate}

\section{Case Study in Catalysis}
\label{Case}

A catalyst is some chemical that is not consumed in the process of a chemical reaction. Using a catalyst in a chemical reaction usually allows said reaction to take place faster and allows for more moderate reaction conditions. Catalysis-based chemical synthesis is applied at roughly 90\% of chemical processes in chemical industry. The scientific domain of catalysis is highly interconnected to other sciences and thus spans over many topics from material sciences to process design \cite{C1, C2}.

\subsection{Used Data}
The texts that have been used for fine-tuning BERT, have been taken from scientific papers in catalysis. These articles have been by PowerShell script extracted to Microsoft Word documents. Thanks to its engine, paragraphs and titles are marked properly, so paragraphs with relevant texts have been extracted and with BERT embedding prepared for classification.

We conduct our experiments on a set of five ontologies from the chemical domain (Table~\ref{CountOntoLabels}) gathered within the \href{https://nfdi4cat.org/en/services/ontology-collection/}{NFDI4Cat project} \cite{wulf21unified}. The ontologies NCIT, CHMO and Allotrope have a close connection to the chemical domain. However, according to their names, the chemical entities of biological interest (CHEBI) and the system biology ontology (SBO) are expected to be further away from the chemical domain. This does not hold necessarily true for the CHEBI as it describes a plethora of chemical entities, also relevant in the chemical and not only biological domain. The SBO was selected as it contains some general laboratory and computational contexts. It also can be seen as some kind of a test, whether the tools used can also identify ontologies not fitting to the text content.

\begin{table}[hbtp]
	\caption{Types and counts of labels in the used OWL files}
	\label{CountOntoLabels}
	\begin{center}
	\begin{tabular}{|c|c|r|}
\hline
\textbf{\begin{tabular}[c]{@{}c@{}}Ontology\\ name\end{tabular}} & \textbf{XML classes} & \multicolumn{1}{c|}{\textbf{\begin{tabular}[c]{@{}c@{}}Number \\of classes\end{tabular}}} \\
\hline
Allotrope & \begin{tabular}[c]{@{}c@{}}Literal\\ rdfs:comment\\ rdfs:label\end{tabular} & 2773 \\
\hline
NCIT & \begin{tabular}[c]{@{}c@{}}rdfs:comment\\ rdfs:label\end{tabular} & 1169 \\
\hline
SBO & \begin{tabular}[c]{@{}c@{}}Literal\\ rdfs:comment\\ rdfs:label\end{tabular} & 534 \\
\hline
CHEBI & \begin{tabular}[c]{@{}c@{}}obo:IAO\_0000115\\ rdfs:label\end{tabular} & 35067 \\
\hline
CHMO & \begin{tabular}[c]{@{}c@{}}obo:IAO\_0000115\\ rdfs:comment\\ rdfs:label\end{tabular} & 2521 \\ \hline
	\end{tabular}
	\end{center}
\end{table}

Hence, these ontologies are classes to which classifiers assign new parts of text. The data have been divided into training and testing datasets in stratified proportion 1:1. The testing dataset has been divided into 20 disjoint subsets, assuming that disjointness is a sufficient condition for their independence. The ontologies most frequently occurring in the dataset have been under-sampled in order to mitigate overfitting during training.

\subsection{Experimental Setting}
At first, the PDFs were transformed into Microsoft Word using PowerShell scripts. The output files have been processed by a python library for parsing docx files. As a result relevant paragraphs have been extracted for classification according to the most relevant ontology. The paragraphs considered irrelevant, namely acknowledgement, references, titles and too short paragraphs (shorter than 100 letters) have been skipped.

The annotations in the specifications of given ontologies have been extracted using the \href{https://www.crummy.com/software/BeautifulSoup/}{BeautifulSoup} python parser for XML. Extracted paragraphs have also been used for BERT fine-tuning. The chosen version of the BERT was recobo/chemical-bert-uncased from the Huggingface portal \cite{chbert}. Using the fine-tuned BERT, every paragraph has been transformed into a 768-dimensional numeric vector.

The extraction of annotations from OWL files has been performed using a python XML parser. Individual annotations have been again embedded into the 768-dimensional vector space using the fine-tuned BERT.

For the employed classifiers, their implementations in ScikitLearn \cite{sklearn} has been used. The optimal  values of hyperparameters of each classifier were determined using a 5-fold cross-validation applied to a grid-search with the grid values listed in Table~\ref{HyperParametersGridSearch}. In order to mitigate overfitting, training data have been undersampled. Statistic computations have used the scipy, statsmodels and pingouin python libraries.

\begin{table*}[hbtp]
	\caption{Hyperparameters of individual classifiers that were determined through grid-search on combinations of considered values. In the column Selected are values, that have been selected using a random stratified 5-fold cross-validation applied to a grid-search with the considered values}
	\label{HyperParametersGridSearch}
	\begin{center}
\begin{tabular}{|l|l|l|l|}
\hline
\textbf{Classifier} & \textbf{Hyperparameter} & \textbf{Considered values} & \textbf{Selected} \\ \hline
\multirow{5}{*}{\begin{tabular}[c]{@{}l@{}}Random\\ forest\end{tabular}} & maximal depth & \{5, 7, 9, 11\} & 11 \\
\cline{2-4} & criterion & \{entropy, gini\} & gini \\
\cline{2-4} & count of estimators & \{5, 10, 15, 20, 25, 30\} & 20 \\
\cline{2-4} & \begin{tabular}[c]{@{}l@{}}fraction of features\\ used in each split\end{tabular} & \{0.5, 0.7\} & 0.5 \\
\cline{2-4}  & bootstrap samples & \{false, true\} & true \\
\hline
\multirow{3}{*}{\begin{tabular}[c]{@{}l@{}}Support\\ vector\\ machine\end{tabular}} & \begin{tabular}[c]{@{}l@{}}slack trade-off\\ constant (C)\end{tabular} & \{1, 10, 100, 1000\} & 100 \\
\cline{2-4}  & kernel type & \{linear, radial basic\} & radial basic \\ \cline{2-4}  & \begin{tabular}[c]{@{}l@{}}kernel coefficient\\ gamma\end{tabular} & {[}0.001, 0.0001{]} & 0.001 \\
\hline
\multirow{2}{*}{\begin{tabular}[c]{@{}l@{}}Gaussian\\ process\end{tabular}} & kernel & \begin{tabular}[c]{@{}l@{}}\{radial basic, dot product, mattern,\\ rational quadratic, white kernel\}\end{tabular} & matern \\
\cline{2-4}  & random state & \{0, 50\} & unapplicable \\
\hline
\multirow{4}{*}{\begin{tabular}[c]{@{}l@{}}K nearest\\ neighbors\end{tabular}}  & \begin{tabular}[c]{@{}l@{}}number of considered\\ neighbors\end{tabular} & \{1, 5, 9, 13, 17\} & 9 \\
\cline{2-4} & weights & \{uniform, distance\} & distance \\
\cline{2-4} & algorithm & \{auto, ball tree, kd tree, brute\} & auto \\
\cline{2-4} & distance metric exponent & \{1, 2, 3, 4, 5\} & 2 \\
\hline
\multirow{6}{*}{\begin{tabular}[c]{@{}l@{}}Multi-layer\\ perceptron\end{tabular}} & random state & \{0, 1\} & 0 \\
\cline{2-4} & activation function & \{identity, logistic, tanh, relu\} & tanh \\
\cline{2-4} & optimizer & \{lbfgs, sgd, adam\} & adam \\
\cline{2-4} & hidden layer size & \{1, 4, 16, 64\} & 4 \\
\cline{2-4} & \begin{tabular}[c]{@{}l@{}}strength of L2\\ regularization term\end{tabular} & \{0.0001, 0.05\} & 0.05 \\
\cline{2-4} & \begin{tabular}[c]{@{}l@{}}learning rate for\\ weights update\end{tabular} & \{constant, adaptive\} & constant \\
\hline
\end{tabular}
	\end{center}
\end{table*}

\subsection{Comparison of Important Classifiers on Considered Ontologies}
\begin{figure*}[pt]
  \centering
    \includegraphics[width=0.99\textwidth]{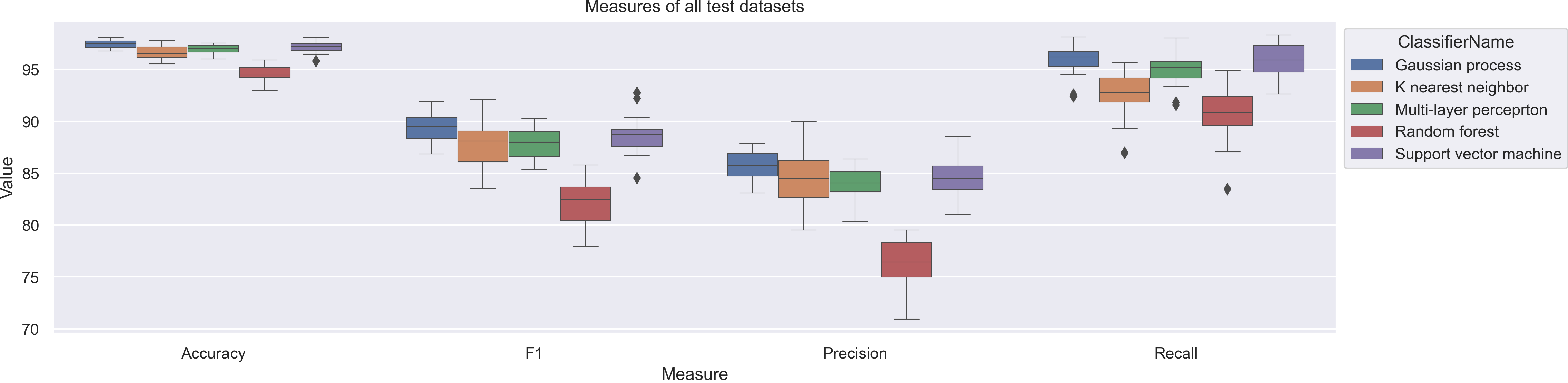}
    \caption{Box plots comparing the distribution of quality measures of the considered classifiers on testing datasets}
    \label{BoxStats}
\end{figure*}

Summary statistics of the predictive accuracy of classifying all 20 testing datasets are in Table \ref{stats}. The table is complemented with boxplots (Figure \ref{BoxStats}), where the following quality measures are presented for each classifier: accuracy, F1 score, precision and recall. The random forest classifier had the worst results of all experiments. Other models had significantly better results. The best accuracy had the Gaussian process, its mean accuaracy was 97.5 \% with very low standard deviation.

\begin{table*}[hbtp]
	\caption{Quality measures of the considered classifiers aggregated over all 20 testing datasets (mean [\%] $\pm$ standard deviation [\%]), where $Accuracy = \frac{TP + TN}{TP + FN + TN + FP}$, $Precision = \frac{TP}{FP + TP}$, $Recall = \frac{TP}{FN + TP}$ and $F1 = 2 \cdot \frac{Precision \cdot Recall}{Precision + Recall}$, using acronyms TruePositive as TP, FalsePositive as FP, TrueNegative as TN and FalseNegative as FN}
	\label{stats}
	\begin{center}
	\begin{tabular}{lrrrr}
\hline
 & \multicolumn{1}{l}{Accuracy} & \multicolumn{1}{l}{F1} & \multicolumn{1}{l}{Precision} & \multicolumn{1}{l}{Recall} \\
\hline
Gaussian process       & $97.46 \pm0.39$ & $89.48 \pm1.38$ & $85.70 \pm1.35$ & $95.88 \pm1.21$ \\
K-nearest neighbor     & $96.66 \pm0.67$ & $87.60 \pm2.41$ & $84.36 \pm2.69$ & $92.73 \pm2.04$ \\
Multi-layer perceptron & $96.99 \pm0.67$ & $87.84 \pm1.59$ & $84.03 \pm1.54$ & $94.97 \pm1.58$ \\
Random forest          & $94.63 \pm0.69$ & $82.00 \pm2.29$ & $76.30 \pm2.34$ & $90.76 \pm2.58$ \\
Support vector machine & $97.16 \pm0.53$ & $88.72 \pm1.85$ & $84.64 \pm1.89$ & $95.85 \pm1.69$ \\ \hline
\end{tabular}
	\end{center}
\end{table*}

\begin{table*}[pt]
	\caption{Comparison of accuracy results on all 20 testing sets with ontology annotations. The values in the table are counts of datasets, in which the model in the row has a higher accuracy compared to the model in the column. If the difference is not significant in the Wilcoxon test then the count is in italic. If the difference is significant, then the higher count is in bold.}
	\label{tests}
	\begin{center}
	\begin{tabular}{lrrrrrr}
\hline
& \multicolumn{1}{l}{\begin{tabular}[c]{@{}l@{}}Random\\ forest\end{tabular}} & \multicolumn{1}{l}{\begin{tabular}[c]{@{}l@{}}Support\\ vector\\ machine\end{tabular}} & \multicolumn{1}{l}{\begin{tabular}[c]{@{}l@{}}Gaussian\\ process\end{tabular}} & \multicolumn{1}{l}{\begin{tabular}[c]{@{}l@{}}K-nearest\\ neighbors\end{tabular}} & \multicolumn{1}{l}{\begin{tabular}[c]{@{}l@{}}Multi-layer\\ perceptron\end{tabular}} & \multicolumn{1}{l}{\begin{tabular}[c]{@{}l@{}}Summary\\ score\end{tabular}} \\
\hline
Random forest & - & 0 & 0 & 0 & 0 & 0 \\
Support vector machine & \textbf{20} & - & 2 & \textbf{15} & \textit{13} & 50 \\
Gaussian process & \textbf{20} & \textbf{16} & - & \textbf{17} & \textbf{19} & 72 \\
K-nearest neighbors & \textbf{20} & 3 & 3 & - & \textit{5} & 31 \\
Multi-layer perceptron & \textbf{20} & \textit{4} & 1 & \textit{14} & - & 39 \\
\hline
\end{tabular}
	\end{center}
\end{table*}

The differences between the considered classifiers were tested for significance by the Friedman test. The basic null hypothesis that the mean accuracy for all 5 classifiers coincides was strongly rejected, with the achieved significance $p = 3.02 \times 10^{-12}$. For the post-hoc analysis, we employed the Wilcoxon signed rank test with two-sided alternative for all 10 pairs of the compared classifiers, because of the inconsistence of the more common mean ranks post-hoc test, as pointed out in \cite{benavoli16should}. For correction to multiple hypotheses testing, we used the Holm method. The results are given in Table \ref{tests}, which shows the highest values for the Support vector machine and Gaussian process classifier.

\subsection{Classification of Scientific Texts with respect to  Relevant Ontologies}
In this experiment classifiers trained in the previous experiments were used. For this experiment, we had no ground truth as to which of the available ontologies is the most relevant for each considered paragraph of text. We employed two collections of scientific papers from the area of catalysis. The small one are papers dealing with the topic of methanation of CO2, it consists of 28 PDFs, from which we have extracted 1~485 relevant paragraphs. The large one is the digital archive of papers (co-)authored by scientists from the Leibniz Institute of Catalysis (with the exception of very few papers with read protection), it consists of 3~450 PDFs, from which we have extracted 144~490 relevant paragraphs. The BERT embeddings of those paragraphs were classified by the five trained classifiers. Every paragraph can be classified into several classes simmultaneously, according to some probability distribution over the available classes. The probability of a particular class in this distribution is considered as confidence of classifying the paragraph into that class.

\begin{figure*}[htp]
  \centering
    \includegraphics[width=0.99\textwidth]{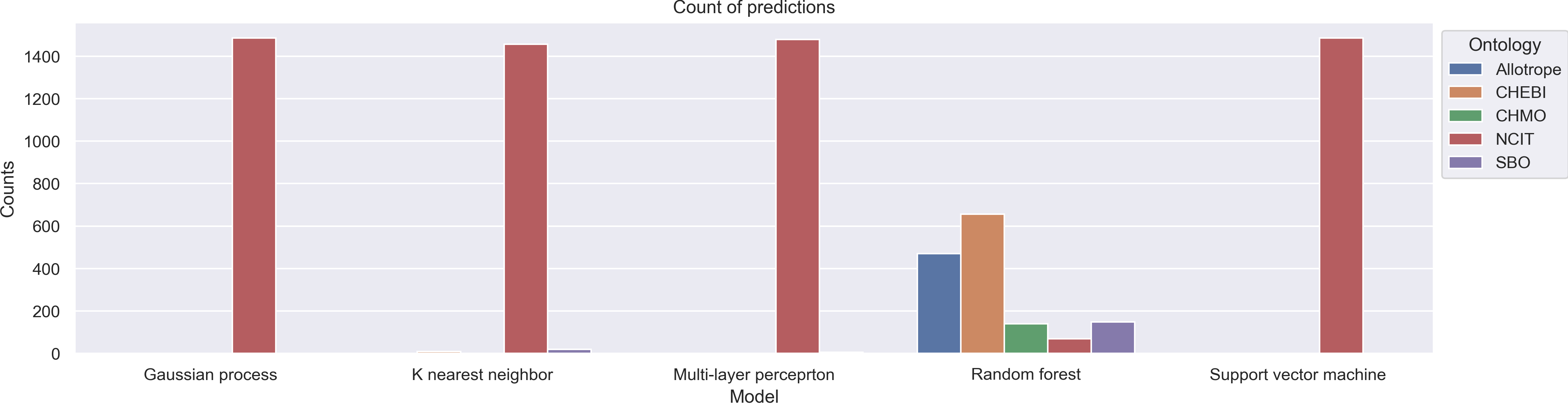}
    \caption{Counts of paragraphs of the small collection of scientific papers predicted according to the highest confidence in the target class}
    \label{CountPredictions}
\end{figure*}

\begin{figure*}[htp]
  \centering
    \includegraphics[width=0.99\textwidth]{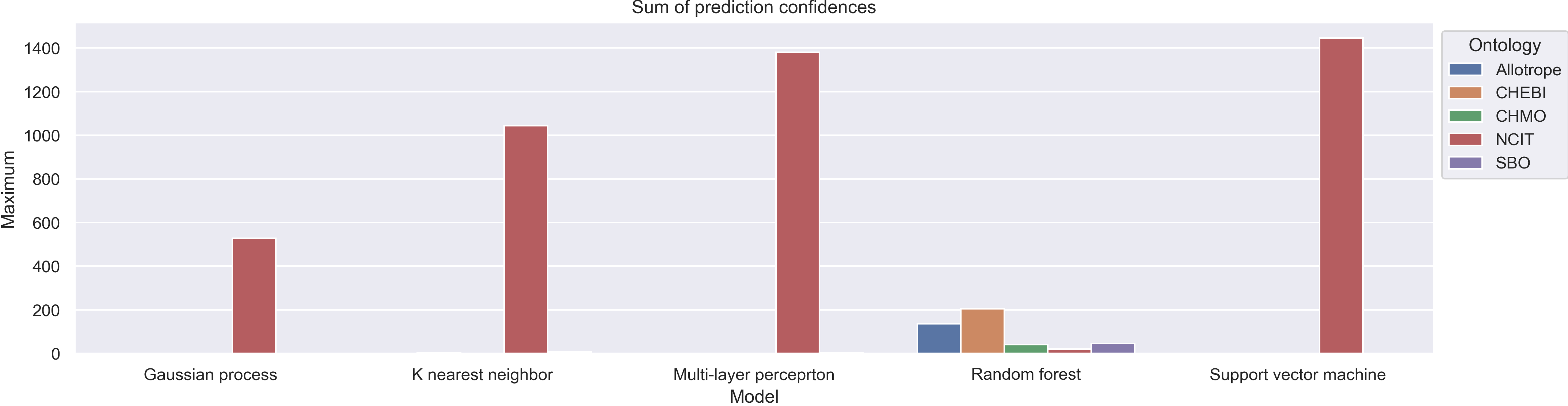}
    \caption{Sum of prediction confidences for the small collection of scientific papers}
    \label{SumConfidencesPredictions}
\end{figure*}

\begin{figure*}[htp]
  \centering
    \includegraphics[width=0.99\textwidth]{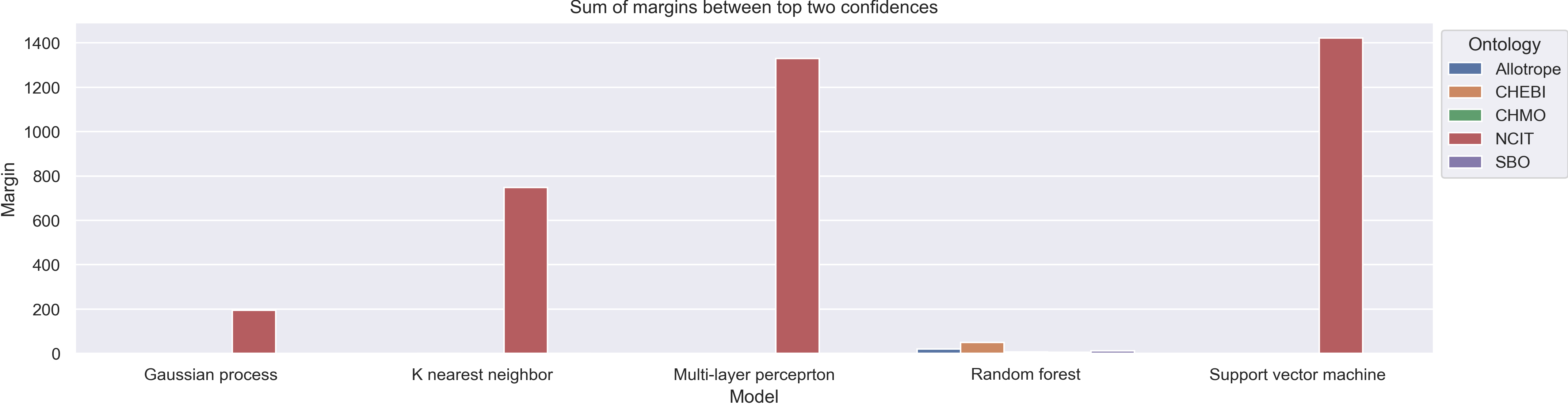}
    \caption{Sum of margins between top two confidences for the small collection of scientific papers}
    \label{MarginTopConfidences}
\end{figure*}

\begin{figure*}[htp]
  \centering
    \includegraphics[width=0.99\textwidth]{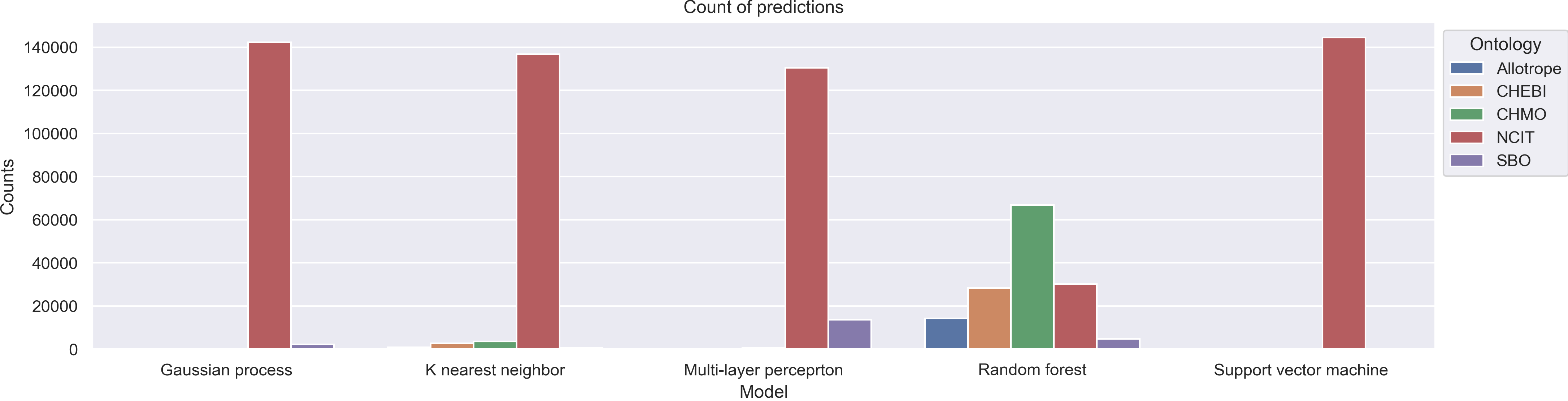}
    \caption{Counts of paragraphs of the large collection of scientific papers predicted according to the highest confidence in the target class}
    \label{CountPredictionsL}
\end{figure*}

\begin{figure*}[htp]
  \centering
    \includegraphics[width=0.99\textwidth]{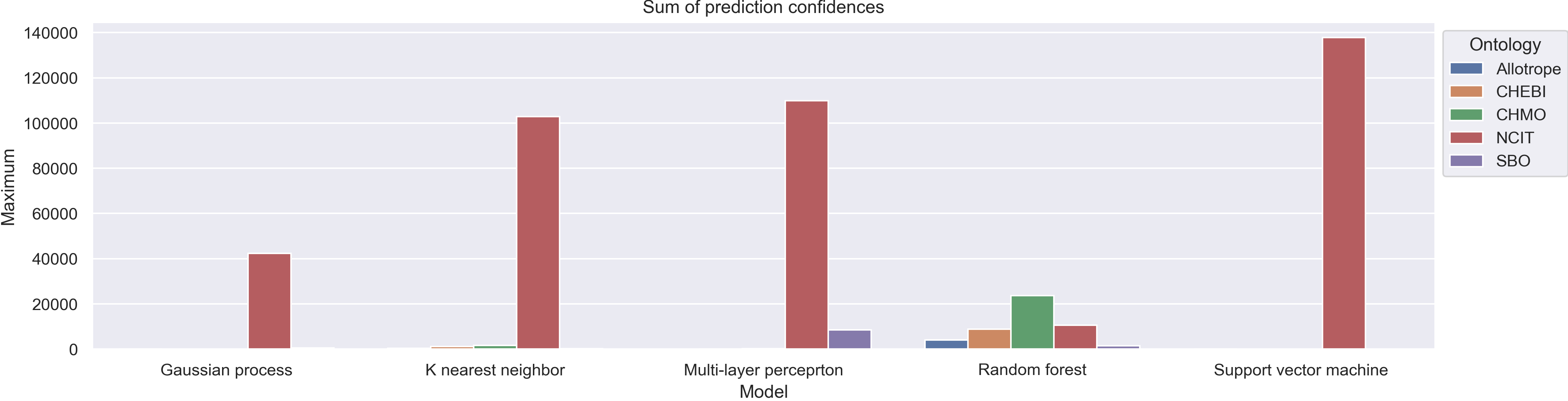}
    \caption{Sum of prediction confidences for the large collection of scientific papers}
    \label{SumConfidencesPredictionsL}
\end{figure*}

\begin{figure*}[htp]
  \centering
    \includegraphics[width=0.99\textwidth]{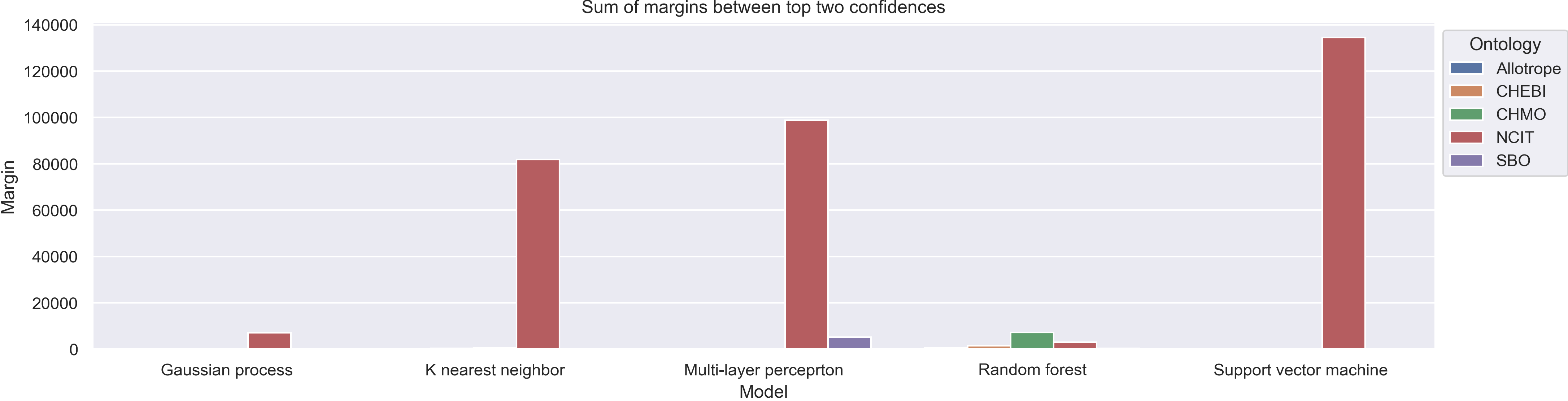}
    \caption{Sum of margins between top two confidences for the large collection of scientific papers}
    \label{MarginTopConfidencesL}
\end{figure*}

\subsubsection{Results for the small dataset}
Figure \ref{CountPredictions} shows how many paragraphs each classifier assigned to each ontology. The Gaussian process, k-nearest neighbor, MLP and SVM assigned almost all paragraphs to the NCIT ontology. The random forest is most uncertain among all classifiers, assigning most paragraphs to the CHEBI ontology, but some pragraphs also to each of the remaining four. Since the contexts of the 28 PDFs fall into the same domain of methanation reactions, it is assumed that most of the paragraphs belong to the same ontology. This seems plausible since the content of the paragraphs is quite similar and focuses on knowledge that may only cover a small part of the five ontologies investigated.

Figure \ref{SumConfidencesPredictions} uses instead of the count of class predictions their confidences. The confidence of the SVM and MLP is very high, whereas that of the Gaussian process and random forest is substantially lower. The k-nearest neighbors classifier has a rather high confidence, too.

In Figure \ref{MarginTopConfidences}, the margin between the  confidence of the predicted ontology and the second highest class confidence is shown. Again, the highest values are achieved by the SVM and MLP, whereas the Gaussian process and random forest have only small margin between the predicted and second most confident ontology, and the k-nearest neighbor has quite a high margin, but not so high as SVM or MLP.

\subsubsection{Results for the large dataset}
Figure \ref{CountPredictionsL} depicts the count of paragraphs from the large dataset that each classifier assigned to each ontology. The Gaussian process, k-nearest neighbor, MLP and SVM assigned almost all paragraphs to the NCIT ontology. The random forest is most uncertain among all classifiers, assigning most paragraphs to the CHMO ontology, but some pragraphs also to each of the remaining four.

Figure \ref{SumConfidencesPredictionsL} using confidences of class predictions shows, that the confidence of the SVM is very high, whereas that of the Gaussian process and random forest is substantially lower. A rather high confidence have also the MLP and the k-nearest neighbors classifier.

In Figure \ref{MarginTopConfidencesL}, the margin between the  confidence of the predicted ontology and the second highest class confidence is shown. Again, the highest values are achieved by the SVM,  whereas the Gaussian process and random forest have only small margin between the predicted and second most confident ontology, and the MLP and k-nearest neighbor have quite a high margin, but not so high as SVM.

\subsubsection{Summary results for both datasets}
The results for both datasets show that the SVM classifier has very high confidences and very high margins between top two confidences. Hence, the results indicate that for a large majority of the unknown scientific texts, the most relevant ontology is NCIT.

\section{Conclusion}
\label{cf}    
This paper provides an insight into the possibility to automatically determine ontologies most relevant to scientific texts. Successfully processing input texts and ontologies often requires a lot of efforts. Here, classifieres have been used in combination with the representation learning by BERT, that may help make this process faster. Our idea has been to use embedding of each paragraph from PDFs as input to classifiers. We used a pretrained BERT that have been fine-tuned using chemical articles. The output embeddings from fine-tuned BERT were used as an input to the classifiers. We have experimented with five different classifiers, in particular random forest, support vector machine, multilayer perceptron, k-nearest neighbors, and Gaussian process. The random forest was not successful, its accuracy was the worst of all models. The best results had Gaussian process and support vector machine.

In a second experiment the considered classifiers have been tested and compared on scientific papers from the domain of catalysis. The ground truth was not known there. The k-nearest neighbor and Gaussian process had very low margin between first and second highest confidence. The highest confidence among all classifiers had the support vector machine. It had also the highest margin among them.

The most serious weakness of this article is the lack of ground truth for the classification of scientific articles, which makes it impossible to evaluate this classification. Therefore, we plan to use methods for reducing the impact of unknown ground truth. Our idea is to use interpolation between annotations using GPT-2 and GPT-3 networks. GPT (Generative Pre-trained Transformer) \cite{GPT} stands for a series of pre-trained language models, which have been trained with a large dataset of textual information and can be applied to deal with specific language-related tasks. BERT, which was trained with Wiki and books data that contains over 3.3 billion tokens, is popular in natural language understanding tasks, e.g., text classification. However, BERT as a masked language model can only learn contextual representation of words but not organize and generate language, which makes it unsuitable for concept generation. On the other hand, GPTs are autoregressive language models that are trained to predict the next token based on all tokens before it.

In future research, it is desirable to try different transformers. We would like to extract knowledge from ANNs in the context of learning. The main direction of our research is extending and integrating ontologies. We plan to use also graph neural networks to incorporate them into representation learning.

\subsection*{Acknowledgement}

The research reported in this paper has been supported by the  German Research Foundation (DFG) funded projects NFDI2/12020 and 467401796, and by the Grant Agency of the Czech Technical University in Prague, grant No. SGS20/208/OHK3/3T/18.

\balance

%
%

\end{document}